\documentclass[runningheads]{llncs}

\usepackage{styling/eccv}

\usepackage{styling/eccvabbrv}

\usepackage{graphicx}
\usepackage{booktabs}
\usepackage{amsmath}
\usepackage{amssymb}
\usepackage{mathtools}
\usepackage{ulem}
\usepackage{dsfont}
\usepackage{multirow}
\usepackage{caption}
\usepackage{dsfont} % provides double-stroke \mathds

\usepackage[accsupp]{axessibility}  % Improves PDF readability for those with disabilities.

\usepackage{hyperref}

\usepackage{orcidlink}

\renewcommand{\hat}{\widehat}
\renewcommand{\bar}{\overline}
\newcommand{\defn}{\triangleq}

\renewcommand{\vec}[1]{\ensuremath{\boldsymbol{#1}}}

\newcommand{\mc}[1]{\ensuremath{\mathcal{#1}}}

\newcommand{\Real}{{\mathbb{R}}}
\newcommand{\Complex}{{\mathbb{C}}}

\newcommand{\of}[1]{^{\scriptscriptstyle (#1)}}

\newcommand{\tran}{^{\top}}
\newcommand{\herm}{^\textsf{H}}

\DeclareMathOperator{\E}{E}

\DeclareMathOperator{\cov}{cov}

\DeclareMathOperator{\blkdiag}{blkdiag}

\newcommand{\mmse}{_{\mathsf{mmse}}}

\DeclareMathOperator{\Quantile}{Quantile}
\DeclareMathOperator{\BetaBin}{BetaBin}
\newcommand{\test}{_{\mathsf{test}}}
\renewcommand{\cal}{_{\mathsf{cal}}}
\newcommand{\round}[1]{^{[#1]}}
\renewcommand{\eqref}[1]{(\ref{eq:#1})}
\DeclareMathOperator{\MIL}{MIL}
\DeclareMathOperator{\EC}{EC}
\DeclareMathOperator{\CE}{CE}

\begin{document}

%\title{Uncertainty Quantification in Inverse Problems via Conformal Bounds on the Performance of Downstream Tasks}
%\title{Downstream Task Uncertainty Quantification via Conformal Bounds in Inverse Problems}
\title{Task-Driven Uncertainty Quantification in Inverse Problems via Conformal Prediction}

\titlerunning{Task-Driven UQ via Conformal Prediction}

\author{Jeffrey Wen \inst{1}\orcidlink{0000-0003-3001-4086} \and
Rizwan Ahmad \inst{1}\orcidlink{0000-0002-5917-3788} \and
Philip Schniter \inst{1}\orcidlink{0000-0003-0939-7545}}

\authorrunning{J.~Wen et al.}

\institute{The Ohio State University, Columbus OH 43210, USA \\
\email{wen.254@osu.edu, rizwan.ahmad@osumc.edu, schniter.1@osu.edu} \\
}

\maketitle

\begin{abstract}
In imaging inverse problems, one seeks to recover an image from missing/corrupted measurements.
Because such problems are ill-posed, there is great motivation to quantify the uncertainty induced by the measurement-and-recovery process.
Motivated by applications where the recovered image is used for a downstream task, such as soft-output classification, we propose a task-centered approach to uncertainty quantification.
In particular, we use conformal prediction to construct an interval that is guaranteed to contain the task output from the true image up to a user-specified probability, and we use the width of that interval to quantify the uncertainty contributed by measurement-and-recovery.
For posterior-sampling-based image recovery, we construct locally adaptive prediction intervals. Furthermore, we propose to collect measurements over multiple rounds, stopping as soon as the task uncertainty falls below an acceptable level. 
We demonstrate our methodology on accelerated magnetic resonance imaging (MRI): 
\href{https://github.com/jwen307/TaskUQ}{https://github.com/jwen307/TaskUQ}.

%\textr{LNCS guidelines indicate it should be at least 70 and at most 150 words.}
\keywords{Inverse Problems \and Uncertainty Quantification \and Conformal Prediction \and Posterior Sampling \and MRI}
\end{abstract}

\section{Introduction} \label{sec:intro}

In imaging inverse problems, one seeks to recover an image $x$ from measurements $y=h(x)$ that mask, distort, and/or corrupt $x$ with noise \cite{Arridge:AN:19}. 
%arise in a variety of fields from remote sensing \cite{Efremenko:21} to medical imaging \cite{Knoll:SPM:20, Johnson:Radiology:23}. 
Linear inverse problems, where
$y = Ax + \epsilon$ with noise $\epsilon$ and known forward operator $A$,
include deblurring, super-resolution, inpainting, colorization, computed tomography (CT), and magnetic resonance imaging (MRI) \cite{Hammernik:SPM:23}. 
Non-linear inverse problems include phase-retrieval, de-quantization, and image-to-image translation.
%The forward process can be non-linear as in the case for phase-retrieval \cite{Dong:SPM:23} or linear as in super-resolution \cite{Chauhan:IA:23} and magnetic resonance imaging (MRI) \cite{Plewes:JMR:12}.
%For linear inverse problems, the relationship between measurement and the true image is more easily recognized as 
%\begin{equation}
%    y = Ax + \epsilon
%    \label{eq:linear_inverse}
%\end{equation}
%where $A$ is a linear operator and $\epsilon$ is the measurement noise. 
These problems are generally ill-posed, in that it is impossible to perfectly infer $x$ from $y$.

Most image recovery methods provide a single ``point estimate'' $\hat{x}$ from measurement $y$ \cite{Arridge:AN:19}. 
%\cite{Lustig:MRM:07, Bora:ICML:17, Ulyanov:CVPR:18, Eo:MRM:18} . 
%Iterative methods like compressed sensing \cite{Lustig:MRM:07, Bora:ICML:17}, ``plug-and-play'' \cite{Ahmad:SPM:20}, and deep image prior \cite{Ulyanov:CVPR:18} define or learn a prior on the space of realistic images and use this prior as a regularizer in the reconstruction process.
%Other works have proposed end-to-end recovery networks \cite{Eo:MRM:18} that provide a reconstruction with a single pass through a deep neural network. 
From $\hat{x}$ alone, it is difficult to determine accuracy with respect to the true $x$.
That is, $\hat{x}$ does not quantify the uncertainty \cite{Abdar:IF:21} in measurement-and-reconstruction that arises from the ill-posed nature of the inverse problem. 
This is problematic in safety-critical applications like medical imaging, where hallucinations or degradations in $\hat{x}$ can result in costly misdiagnoses \cite{Banerji:NM:23}. 
%Olsson:NCOMM:22, Lu:MICCAI:22
%Unfortunately, this limitation constrains the trustworthiness of the reconstructions and ability to perform informed downstream tasks.

Several approaches have been proposed to provide uncertainty quantification (UQ) within the image recovery process.
One approach is to utilize Bayesian Neural Networks (BNNs), which treat the reconstruction network parameters as random variables \cite{Kendall:NIPS:17, Xue:Optica:19, Barbano:ICPR:21, Ekmekci:TCI:22, Narnhofer:TMI:22}. 
This allows one to quantify epistemic (i.e., model) uncertainty by measuring the variation over reconstructions generated by different draws from the parameter distribution. 
Posterior sampling methods \cite{Durmus:JIS:18, Adler:18, Ardizzone:ICLR:19, Edupuganti:TMI:20, Jalal:NIPS:21, Chung:ICLR:23} 
%Lugmayr:ECCV:20, Kawar:NIPS:22,
instead draw many samples from the distribution $p(x|y)$ of plausible $x$ given $y$, known as the posterior, and aim to quantify the uncertainty that the measurement process imposes on $x$ (i.e., aleatoric uncertainty).
%Combining these approaches, Ekmekci and Cetin \cite{Ekmekci:NIPSW:23} train an ensemble of posterior samplers to quantify the uncertainty over different models and different samples from the posterior. 
It's possible to combine BNNs with posterior sampling, as in \cite{Ekmekci:NIPSW:23}.

Although the samples generated by BNNs and posterior-sampling methods could be used in many ways, they are most commonly used to compute pixel-wise uncertainty images or ``maps.''
To compute pixel-wise uncertainty intervals with statistical guarantees, conformal prediction can be used \cite{Angelopoulos:ICML:22, Horwitz:22, Teneggi:ICML:23, Kutiel:ICLR:23}.
%Beyond sampling procedures, methods based on risk-controlling prediction sets (RCPS) \cite{Angelopoulos:ICML:22, Teneggi:ICML:23, Kutiel:ICLR:23} take a conformal prediction \cite{Angelopoulos:FTML:23, Vovk:Book:05, Lei:JRSS:14, Tibshirani:23} approach and generate pixel-wise intervals bounding the true pixel values with statistical guarantees. 
Still, the value of these pixel-wise uncertainty maps is not clear.
%While these methods are able to provide pixel-wise uncertainty maps, there still remains a gap in the literature for how to best distill the information for real-world use. 
For example, when recovering images, we are usually concerned about many-pixel visual structures (e.g., lesions in MRI, hallucinations) that single-pixel statistics say little about.
Secondly, uncertainty maps are not easy to interpret.
They often convey little beyond the notion that there is less pixel-wise uncertainty in smooth regions as compared to near edges (see, e.g., \cref{fig:mri_intervals}).
Lastly, it's not clear how pixel-wise uncertainty relates to the overall imaging goal, which is often task-oriented, such as detecting whether a tumor is present or not. 
%Thus, we seek a ``higher level'' UQ beyond pixels.

One approach to assess multi-pixel uncertainty is Bayesian Uncertainty Quantification by Optimization (BUQO) \cite{Tang:23}, 
%Repetti:EUSIPCO:18
which aims to test whether a particular ``structure of interest'' in the maximum a-posteriori (MAP) reconstruction is truly present. % or simply a reconstruction artifact. 
However, inpainting is used to hypothesize what the image would look like without the structure, the correctness of which is difficult to guarantee.

\begin{figure}[t]
    \centering
    \includegraphics[width=1.0\linewidth]{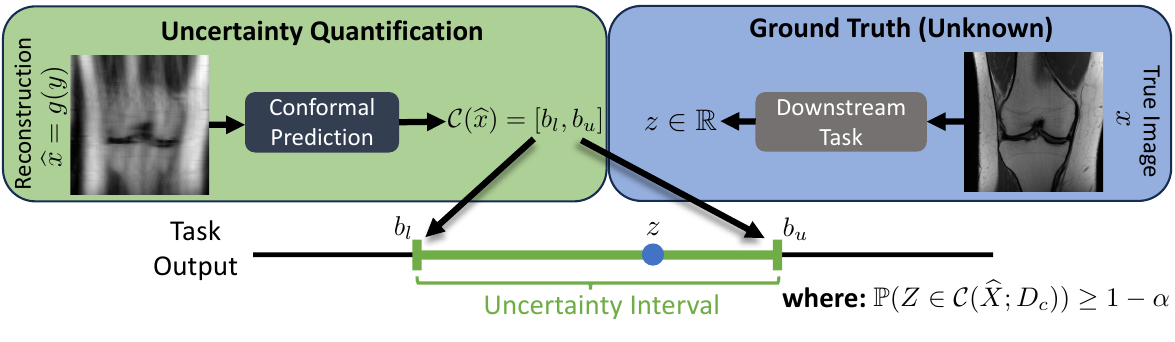}
    \caption{High-level overview of our approach:
    For true image $x$, measurement $y=h(x)$, recovery $\hat{x}=g(y)$, and task output $\hat{z}=f(\hat{x})$, we use conformal prediction to construct an interval $\mc{C}(\hat{x};d\cal)\subset\Real$ that is guaranteed to contain the true task output $z=f(x)$ in the sense that $\mathbb{P}(Z\in\mc{C}(\hat{X};D\cal))\geq 1-\alpha$ for some chosen error-rate $\alpha$.}
    \label{fig:goal}
\end{figure}

In this work, we propose a novel UQ framework for imaging inverse problems that aims to provide a more impactful measure of uncertainty. 
%Our method explicitly defines the uncertainty in a downstream task (i.e. classification, regression) contributed by the measurement process. 
%In particular, our framework quantifies how similar a scalar-output downstream task can be performed by a deep neural network (DNN) given only measurements relative to being given the full ground truth image.
%To do this, we can use any reconstruction model and any DNN trained on ground truth images for the downstream task.
%We take, as a reference point, the network's output when presented with the ground truth image.
%When fed with an imperfectly recovered image, we expect that the downstream output will deviate from the reference, but we would like to guarantee that it is close.
In particular, we aim to quantify to what extent a downstream task behaves differently when supplied with the reconstructed image versus the true image.
Our framework supports any measurement-and-reconstruction procedure and any downstream task that outputs a real-valued scalar.
%As a running example, we consider the soft-output classification task in accelerated magnetic resonance imaging.
%Using conformal prediction, we propose to construct---for a given measurement vector---a downstream-output interval that is guaranteed to contain the ground truth reference value with a user-specified probability.
Our contributions are as follows.
\begin{enumerate}
\item
We propose to construct, using conformal prediction, an interval in the task-output space that is guaranteed to contain the true task output up to a user-specified probability. 
The prediction interval width provides a natural way to quantify the uncertainty that measurement-and-reconstruction contributes to the downstream task output.  
%By providing a single uncertainty value, our framework provides an quick, efficient way to evaluate if the downstream task can be reliability completed using the reconstructions. 
(See \cref{fig:goal}.)
\item
For posterior-sampling-based image reconstruction, we propose to construct adaptive uncertainty intervals that shrink when the measurements offer more certainty about the true output of the downstream task.
(See \cref{fig:method_overview}.)
\item
We propose a multi-round acquisition protocol whereby measurements are accumulated until the task uncertainty is acceptably low.
\item
We demonstrate our approach on accelerated MRI with the task of soft-output-classifying a meniscus tear. Several conformal predictors are evaluated and compared.
\end{enumerate}

\section{Background} \label{sec:background}

Conformal prediction \cite{Vovk:Book:05, Angelopoulos:FTML:23} %Lei:JRSS:14, Tibshirani:23 
is a framework for generating uncertainty sets with prescribed statistical guarantees.
Notably, it can be applied to any black-box predictor without making any distributional assumptions about the data. 
%This makes it extremely versatile for a broad range of applications. 

We now explain the basics of conformal prediction or, more precisely, the common variant known as split conformal prediction \cite{Papadopoulos:ECML:02,Lei:JASA:18}.
Say that we have a black-box model $f:\mathcal{X}\rightarrow\mathcal{Z}$ that predicts a target $z \in \mathcal{Z}$ from features $x \in \mathcal{X}$.
%We focus on $\mathcal{Z}\subset\Real$, but categorical $\mathcal{Z}$ can also be supported.
Say that we also have a calibration dataset $d\cal \defn \{(x_i, z_i)\}_{i=1}^n$ that was unseen when training $f$, as well as a test feature $x\test$ and an unknown test target $z\test$.
In split conformal prediction, we use the calibration data to construct a prediction set $\mathcal{C}(x\test;d\cal)\subset 2^{\mc{Z}}$ that provides the ``marginal coverage'' \cite{Lei:JRSS:14} guarantee
\begin{equation}
\mathbb{P}\big(Z\test \in \mathcal{C}(X\test; D\cal)\big) \geq 1 - \alpha 
\label{eq:coverage},
\end{equation}
where $\alpha \in [0,1]$ is a user-specified error rate. 
In \eqref{coverage} and elsewhere, we use capital letters to denote random variables and lower-case to denote their realizations.
Thus \eqref{coverage} can be interpreted as follows:
When averaged over random calibration data $D\cal$ and test data $(X\test,Z\test)$, the set $\mathcal{C}(X\test;D\cal)$ is guaranteed to contain the correct target $Z\test$ with probability no less than $1-\alpha$.
Although we would prefer a ``conditional coverage'' guarantee of the form $\mathbb{P}\big(Z\test \in \mathcal{C}(X\test;D\cal)\big|X\test=x\test\big) \geq 1 - \alpha$, this is generally impossible to achieve \cite{Vovk:ACML:12,Lei:JRSS:14}.

We now describe the standard recipe for constructing a prediction set\linebreak $\mathcal{C}(x\test;d\cal)$. 
First one chooses a nonconformity score $s(x,z;f)\in\Real$ that assigns higher values to worse predictions. 
Then one computes the empirical quantile
\begin{equation}
\hat{q} \defn \Quantile\Big( \tfrac{\lceil (1-\alpha)(n+1) \rceil}{n}; s_1,\dots,s_n \Big)
\label{eq:empirical_quantile}
\end{equation}
from the calibration scores $s_i=s(x_i,z_i;f)$.
Finally one constructs 
\begin{equation}
\mc{C}(x\test;d\cal) = \{z: s(x\test,z;f) \leq \hat{q}\}
\label{eq:prediction_set} .
\end{equation}
%where, with a slight abuse of notation, we summarized the calibration set $d\cal$ in $\mc{C}(\hat{x}\test;f,d\cal)$ by its score-quantile $\hat{q}$.
Under these choices, it can be proven \cite{Vovk:ICML:99,Lei:JASA:18} that the marginal coverage guarantee \eqref{coverage} holds when $(X_1,Z_1),\dots,(X_n,Z_n),(X\test,Z\test)$ are i.i.d., and even under the weaker condition that they are exchangeable \cite{Vovk:Book:05}.
%Since it's clear from \eqref{prediction_set} that the role of the calibration set $d\cal$ is entirely summarized by the score-quantile $\hat{q}$, we write ``$\mc{C}(x\test;f,\hat{q})$'' in place of ``$\mc{C}(x\test;f,d\cal)$'' in the sequel.

There are many ways to construct the nonconformity score $s(x,z;f)$.
For real-valued targets $z$, the simplest choice would be the absolute residual
\begin{equation}
s(x,z; f) = |z - f(x)|
\quad\Rightarrow\quad
\mathcal{C}(x;d\cal) = \big[f(x)-\hat{q},f(x)+\hat{q}\big]
\label{eq:absolute_residual} ,
\end{equation}
which gives an $x$-invariant interval length of $|\mathcal{C}(x;d\cal)|=2\hat{q}$.
%An alternative is to use quantile regression \cite{Koenker:ECON:78}, where $f(x;\gamma)$ approximates the $\gamma$-quantile of $Z\test|X\test=x$.
%Then by using the score \cite{Romano:NIPS:19}
%\begin{equation}
%s(x,z; f) = \max\Big\{ f\Big(x;\frac{\alpha}{2}\Big)-z, z-f\Big(x;1-\frac{\alpha}{2}\Big)\Big\}
%\label{eq:s_quantile} ,
%\end{equation}
%we obtain a prediction interval whose width locally adapts to $x$:
%\begin{equation}
%\mathcal{C}(x;f,\hat{q}) = \Big[f\Big(x;\frac{\alpha}{2}\Big)-\hat{q},f\Big(x;1-\frac{\alpha}{2}\Big)+\hat{q}\Big]
%\label{eq:C_quantile} .
%\end{equation}
We will discuss a few other choices in the sequel.
For more on conformal prediction, we suggest the excellent overviews \cite{Vovk:Book:05, Angelopoulos:FTML:23}.

\section{Proposed Method}\label{sec:method}

Suppose that we collect measurements $y=h(x)$ of the true image $x$,
from which we compute an image recovery $\hat{x}=g(y)$.
We would ideally like that $\hat{x}=x$, but this is impossible to guarantee with an ill-posed inverse problem.
Although there are many ways to quantify the difference between $\hat{x}$ and $x$ (e.g., PSNR, SSIM \cite{Wang:TIP:04}, LPIPS \cite{Zhang:CVPR:18}, DISTS \cite{Ding:TPAMI:20}), we will instead assume that we are primarily interested in using $\hat{x}$ for some downstream task $f(\hat{x})\in\Real$.
As a running example, we consider $x$ to be a medical image, $y$ to be accelerated MRI measurements, and $f(\cdot)\in[0,1]$ to be the soft output of a classifier that aims to detect the presence or absence of a pathology. 
For example, when $f(\hat{x})=0.7$, the classifier believes that there is a 70\% chance that the pathology exists. 

When image recovery is imperfect (i.e., $\hat{x}\neq x$), we expect the task output to also be imperfect, in the sense that $\hat{z}=f(\hat{x})\neq f(x)=z$. 
%In the ideal case that measurement-and-reconstruction $g\circ h$ was lossless, the computed task output $\hat{z}\defn f(\hat{x})$ would perfectly predict the true task output $z\defn f(x)$.
%In practice, however, $\hat{z}$ is an imperfect estimate of the true $z$.  
We are thus strongly motivated to understand how close $\hat{z}$ is to the true $z$ or, even better, to construct a prediction interval $\mc{C}(\hat{x})\subset\Real$ that contains the true $z$ with some guarantee. 
The interval width $|\mc{C}(\hat{x})|$ would then quantify the uncertainty that the measurement-and-reconstruction process contributes to predicting the true task output $z$.

We emphasize that our approach makes \textit{no assumptions about the task} $f(\cdot)$ beyond it producing a real number.
For example, if $f(\cdot)$ is a soft-output classifier, we do not assume that it is accurate or even calibrated \cite{Guo:ICML:17}.
Likewise, our approach does not aim to assess the uncertainty implicit in the task, but rather the \textit{additional uncertainty that measurement-and-reconstruction contributes to the task}.
For a soft-output classifier, a (true) output of $z=f(x)=0.7$ would express considerable uncertainty about the presence of a pathology in $x$.
But if the true $z$ could be perfectly predicted from $\hat{x}$, then the measurement-and-reconstruction process would bring no \textit{additional} uncertainty.

To construct the interval $\mc{C}(\hat{x})$, we use conformal prediction.
Adapting the methodology from \cref{sec:background} to the current setting, we use a calibration set $d\cal=\{(\hat{x}_i,z_i)\}_{i=1}^n$ of (recovered-image, true-task-output) pairs, and we expect to satisfy the marginal coverage guarantee
\begin{equation}
\mathbb{P}\big(Z \in \mathcal{C}(\hat{X}; D\cal)\big) \geq 1 - \alpha 
\label{eq:mri_coverage} 
\end{equation}
when $(\hat{X}_1,Z_1),\dots,(\hat{X}_n,Z_n),(\hat{X},Z)$ are exchangeable.
In \eqref{mri_coverage} and in the sequel, we explicitly denote the dependence of $\mc{C}(\hat{x})$ on the calibration data.
To construct the calibration set, we assume access to ground-truth examples $\{x_i\}_{i=1}^n$, a measurement model $h(\cdot)$, a reconstruction model $g(\cdot)$, and a task function $f(\cdot)$, from which we can construct $y_i=h(x_i)$, $\hat{x}_i=g(y_i)$, and $z_i=f(x_i)$ for $i=1,\dots,n$.

In some cases we may instead have access to a posterior-sampling-based image reconstruction model that generates $p$ recoveries $\{\hat{x}_i\of{j}\}_{j=1}^p$ from every measurement $y_i$ via $\hat{x}_i\of{j}=g(y_i,v_i\of{j})$, where $\{v_i\of{j}\}_{j=1}^p$ are i.i.d.\ code vectors and, typically, $v_i\of{j}\sim\mc{N}(0,I)$.
In this case, the prediction interval becomes $\mc{C}(\{\hat{x}_i\of{j}\}_{j=1}^p;d\cal)$.
As we will see, posterior sampling facilitates locally adaptive prediction sets.

\begin{figure}[t]
    \centering
    \includegraphics[width=1.0\linewidth]{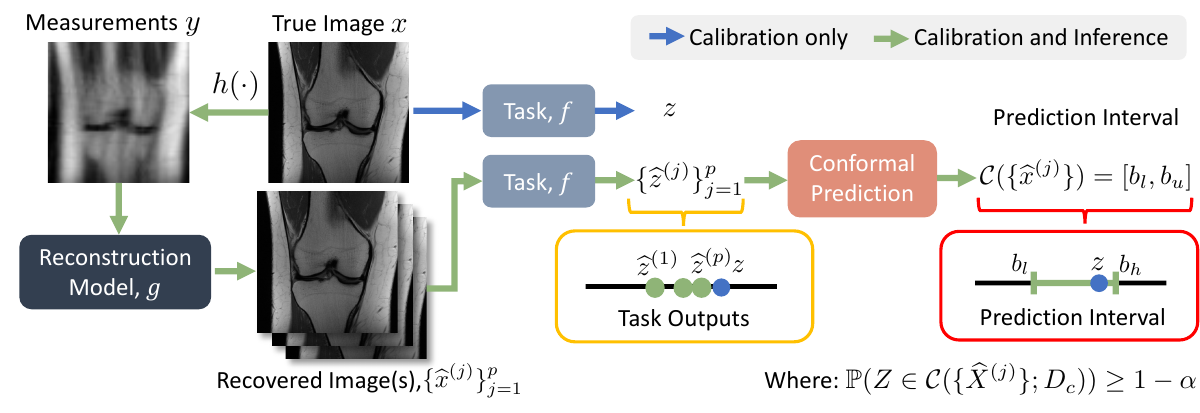}
    \caption{Detailed overview of our approach: 
    For true image $x$, measurement $y=h(x)$, reconstructions $\{\hat{x}\of{j}\}_{j=1}^p$, and task outputs $\hat{z}\of{j}=f(\hat{x}\of{j})$, we use conformal prediction with a calibration set $d\cal=\{(\{\hat{x}_i\of{j}\}_{j=1}^p,z_i)\}_{i=1}^n$ to construct an interval $\mc{C}(\{\hat{x}\of{j}\};d\cal)=[b_l,b_h]$ that is guaranteed to contain the true task output $z=f(x)$ in the sense that  $\mathbb{P}(Z\in\mc{C}(\{\hat{X}\of{j}\};D\cal))\geq 1-\alpha$ for some chosen error-rate $\alpha$.}
    \label{fig:method_overview}
\end{figure}

Next we describe different ways to construct the prediction intervals $\mc{C}(\hat{x};d\cal)$ and $\mc{C}(\{\hat{x}_i\of{j}\}_{j=1}^p;d\cal)$, and later we describe a multi-round measurement protocol that exploits locally adaptive prediction intervals.
See \cref{fig:method_overview} for a detailed overview of our approach.

%To construct the intervals, we use conformal prediction. We assume we have access to measurements $Y_{i} \sim \mathcal{Y}$ and corresponding true images $X_{i} \sim \mathcal{X}$ that are distinct from the training data of any network used. By passing the true images through the downstream network, we obtain reference values $Z_{i}=f(X_{i})$ with which we can construct a calibration set $\mathcal{D}_c = \{(Y_{i}, Z_{i})\}_{i=1}^{n}$ that we will use to tune our conformal intervals. For the inverse problem, we also assume we have a reconstruction network $g$ that is trained to provide estimated reconstructions $\hat{x}=g(y)$ from measurements. Note that $g$ can be any generic point estimate or posterior sampling reconstruction network. We can define any nonconformity score $s(y, z; f, g)$ and compute the empirical quantile $\hat{q}$ as in \cref{eq:empirical_quantile} using $\mathcal{D}_c$. Finally, we construct conformal intervals $\mathcal{C}(\hat{x}; f, g, \hat{q})$ such that
%\begin{equation}
%    \mathbb{P}(Z\test \in \mathcal{C}(\hat{X}\test; f,g, \hat{Q})) \geq 1 - \alpha 
%    \label{eq:mri_coverage}
%\end{equation}
%for an exchangeable test sample $(Y\test, Z\test)$ and user-defined error rate $\alpha$. An overview of our method can be visualized in \cref{fig:method_overview}

%Using this framework, we can apply a range of conformal methods by simply changing the nonconformity score. We propose three to demonstrate the generality. For simplicity, we drop the dependence on $f$, $g$, and $\hat{q}$ in the proceeding notation.

\subsection{Method 1: Absolute Residuals (AR)} \label{sec:AR}

We first consider the case where image recovery yields a point-estimate $\hat{x}=g(y)$ of the true $x$.
As described in \cref{sec:background}, a simple way to construct a nonconformity score is through the absolute residual (recall \eqref{absolute_residual})
%Suppose our reconstruction model $g$ is a point estimate method that gives only a single estimate $\hat{x}$ for a given measurement $y$. We can establish a conformal interval using the absolute residual as in \cref{eq:absolute_residual}. We use the nonconformity score function 
\begin{equation}
s(\hat{x}, z; f) = |z - f(\hat{x})|
\label{eq:s_AR}.
\end{equation}
Evaluating this score on the calibration set $d\cal$ gives $\{s_i\}_{i=1}^n$, whose empirical quantile $\hat{q}$ can be computed as in \eqref{empirical_quantile} and used to construct the prediction interval
%Computing the empirical quantile of the calibration set using \eqref{empirical_quantile}, we can find the conformal interval
\begin{equation}
\mathcal{C}(\hat{x};d\cal) = \big[f(\hat{x}) - \hat{q}, f(\hat{x}) + \hat{q}\big]
\label{eq:C_AR} ,
\end{equation}
which then provides the marginal coverage property \eqref{mri_coverage} \cite{Lei:JASA:18}.

Note that, with this choice of score, the interval width $|\mc{C}(\hat{x};d\cal)|=2\hat{q}$ varies with the calibration set $d\cal$ but not with $\hat{x}$.
Thus, for a fixed $d\cal$, the score \eqref{s_AR} provides no way to tell whether one $\hat{x}$ will yield more task-output uncertainty than a different $\hat{x}$. 
%Imagine two scenarios, one where $\hat{x}$ tends to be a better estimate of $x$ and another where $\hat{x}$ tends to be a worse estimate, and suppose that a separate calibration set was collected for each. 
%Then we would expect the first scenario to yield a smaller prediction interval than the latter, implying less uncertainty in predicting the task output $z$.
%But for a fixed choice of calibration set $d\cal$, the score \eqref{s_AR} provides no way to tell whether one realization $\hat{x}$ will yield more task-output uncertainty than a different $\hat{x}$. 

%While this method does give uncertainty estimates, the intervals are fixed for all samples once the calibration process is completed. Thus, the amount of uncertainty is based on the number of measurements.This could provide insight into how many measurements need to be collected to achieve a chosen uncertainty level on average but limits the applicability to multi-round measurement procedures. Ideally, we would like to have intervals that adapt with the difficulty of the input. In other words, if the measurements contain more evidence for a consistent downstream output, then the interval should be smaller than for measurements with inconclusive reconstructions. This property is known as local adaptivity \cite{Lei:JASA:18, Romano:NIPS:19, Romano:NIPS:20, Angelopoulos:ICLR:20}. The next two proposed methods facilitate local adaptivity by utilizing posterior sampling methods.

\subsection{Method 2: Locally-Weighted Residuals (LWR)} \label{sec:LWR}

We now consider the case where we have a posterior-sampling-based recovery method that yields $p$ recoveries $\{\hat{x}\of{j}\}_{j=1}^p$ per measurement $y$.
We make no assumption on how accurate or diverse these $p$ samples are, other than assuming that the corresponding task-outputs $\hat{z}\of{j}=f(\hat{x}\of{j})$ are not all identical.

Suppose that we choose the nonconformity score
\begin{equation}
s(\{\hat{x}\of{j}\},z;f) = \frac{|z - \bar{z}|}{\sigma_z}
\text{~with~}
\begin{cases}
\bar{z} \defn \frac{1}{p} \sum_{j=1}^{p}f(\hat{x}\of{j})\\
\sigma_z \defn \sqrt{\frac{1}{p} \sum_{j=1}^{p} \big(f(\hat{x}\of{j}) - \bar{z}\big)^2}
\end{cases}
\label{eq:s_LWR} ,
\end{equation}
evaluate it on the calibration set $d\cal$ to get scores $\{s_i\}_{i=1}^n$, and compute their empirical quantile $\hat{q}$ as in \eqref{empirical_quantile}.
Then the prediction interval
\begin{equation}
\mathcal{C}(\{\hat{x}\of{j}\};d\cal) 
= \big[\bar{z} - \sigma_z \hat{q}, \bar{z} + \sigma_z \hat{q} \big]
\label{eq:C_LWR}
\end{equation}
of this ``locally weighted residual'' (LWR) method provides the marginal coverage property in \eqref{mri_coverage} \cite{Lei:JASA:18}.

In words, this method first computes (approximate) posterior samples $\hat{z}\of{j}\sim Z|Y=y$, which are then averaged to approximate the conditional mean $\hat{z}\mmse\defn\E(Z|Y=y)\approx\bar{z}$ and square-root conditional covariance $\sqrt{\cov(Z|Y=y)}\approx\sigma_z$.
When exactly computed, the conditional covariance gives a meaningful uncertainty metric on how well the true $Z$ can be estimated from measurements $y$, because $\cov(Z|Y=y)=\E((Z-\hat{z}\mmse)^2|Y=y)$. 
However, the $\sigma_z$ that we compute is merely an approximation.
So, with the aid of the calibration set, $\sigma_z$ is adjusted by the scaling $\hat{q}$ to yield a prediction interval $\big[\bar{z} - \sigma_z \hat{q}, \bar{z} + \sigma_z \hat{q} \big]$ that satisfies the marginal coverage criterion \eqref{mri_coverage}.

Importantly, the interval width $|\mc{C}(\{\hat{x}\of{j}\};d\cal)|$ now varies with $\{\hat{x}\of{j}\}_{j=1}^p$ through $\sigma_z$.
This latter property is known as ``local adaptivity'' \cite{Lei:JASA:18}.

\subsection{Method 3: Conformalized Quantile Regression (CQR)} \label{sec:CQR}

Another popular locally adaptive method is known as conformalized quantile regression (CQR) \cite{Romano:NIPS:19}.
The idea is to construct the nonconformity score using two quantile regressors \cite{Koenker:ECON:78}, one which estimates the $\frac{\alpha}{2}$th quantile of $Z|Y=y$ and the other which estimates the $(1\!-\!\frac{\alpha}{2})$th quantile.

To compute these quantile estimates, we will once again assume access to a posterior-sampling-based recovery method that yields $p$ recoveries $\{\hat{x}\of{j}\}_{j=1}^p$ per measurement $y$.
From the corresponding task-outputs $\hat{z}\of{j}=f(\hat{x}\of{j})$, we compute the empirical quantiles $\hat{z}(\frac{\alpha}{2})$ and $\hat{z}(1-\frac{\alpha}{2})$ using 
\begin{equation}
\hat{z}(\omega) \defn \Quantile\big(\omega;\hat{z}\of{1},\dots,\hat{z}\of{p}\big) .
\end{equation}
From these quantile estimates, we construct the nonconformity score
\begin{equation}
s(\{\hat{x}\of{j}\},z;f)
= \max\big\{\hat{z}(\tfrac{\alpha}{2})-z,z-\hat{z}(1-\tfrac{\alpha}{2})\big\} ,
\end{equation}
evaluate it on the calibration set $d\cal$ to obtain $\{s_i\}_{i=1}^n$, and compute their $\lceil(1-\alpha)(n+1)\rceil/n$-empirical quantile $\hat{q}$ as in \eqref{empirical_quantile}.
Then the prediction interval
\begin{equation}
\mathcal{C}(\{\hat{x}\of{j}\};d\cal) = \big[\hat{z}(\tfrac{\alpha}{2}) - \hat{q}, \hat{z}(1-\tfrac{\alpha}{2}) + \hat{q} \big]
\label{eq:C_CQR}
\end{equation}
provides the marginal coverage in \eqref{mri_coverage} \cite{Romano:NIPS:19}.
Like \eqref{C_LWR}, this interval is locally adaptive.
We will compare these three conformal prediction methods in \cref{sec:experiments}.

%Another popular form of locally adaptive methods is conformalized quantile regression (CQR) \cite{Romano:NIPS:19}. 
%This method trains  and then applies the conformal procedure to find an additive adjustment that allows the quantiles to provide the desired coverage. As opposed to training two quantile regressors, we simply find the $\alpha / 2$ and $1-\alpha/2$ quantiles based on our single downstream network's outputs for the posterior samples. To do this, we define a function to find the $\tau$-level quantile
%\begin{equation}
%    f^{\tau}(y, \tilde{v}) = \Quantile(\tau, f(g(y,v\of{1}))...f(g(y, v\of{p})))
%\end{equation}
%Using this function, we can find an interval between the $\alpha / 2$ and $1-\alpha/2$ quantiles, but again, this interval does not provide any statistical insurances with regard to the ground truth prediction. Thus, we seek to conformalize the interval. We define the new nonconformity score as
%\begin{equation}
%    s(y, \tilde{v}, z) = \max \{ f^{\alpha / 2}(y, \tilde{v}) - z, z - f^{1-\alpha/2}(y, \tilde{v}) \}
%\end{equation}
%and the conformal interval as
%\begin{equation}
%    \mathcal{C}(\hat{x}, \tilde{v}) = [f^{\alpha / 2}(y, \tilde{v}) - \hat{q}, f^{1-\alpha/2}(y, \tilde{v}) + \hat{q} ]
%\end{equation}
%In this context, we can view the empirical quantile $\hat{q}$ as an additive adjustment to the interval defined by the $\alpha / 2$ and $1-\alpha/2$ quantiles, which provides the coverage guarantees of \cref{eq:mri_coverage}. 

\subsection{Multi-Round Measurement Protocol}  \label{sec:protocol}

\begin{figure}[t]
    \centering
    \includegraphics[width=1.0\linewidth]{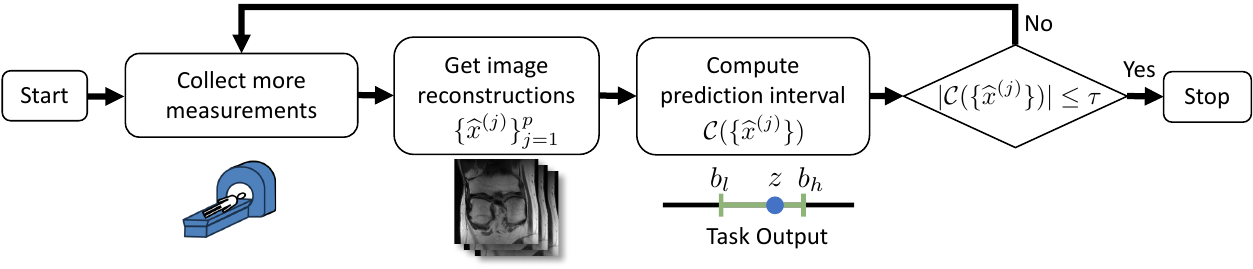}
    \caption{Proposed multi-round measurement protocol. In each round, measurements are collected and reconstructions and conformal intervals are computed. If the length of the interval falls below a user-set threshold $\tau$, the procedure stops. Otherwise, more measurements are collected, and the process repeats until the threshold has been met.}
    \label{fig:multi_round_sampling}
\end{figure}

In many applications, there is a significant cost to collecting a large number of measurements (i.e., acquiring a high-dimensional $y$). 
One example is MRI, the details of which will be discussed in \cref{sec:experiments}.
%In MRI, for example, the cost is measured in time.
%Conventional MRI exams take about 30 minutes, which poses challenges to patient comfort and limits how many patients can be processed per day.
For these applications, we propose to collect measurements over multiple rounds, stopping as soon as the task uncertainty falls below a prescribed level $\tau$.
The goal is to collect the minimal number of measurements that accomplishes the task with probability of at least $1-\alpha$.

Our approach is to use the prediction interval width $|\mc{C}(\{\hat{x}\of{j}\};d\cal)|$ as the metric for task uncertainty.  
This requires the interval to be locally adaptive, as with LWR and CQR above.
The details are as follows.
First, a sequence of $C>1$ nested measurement configurations is chosen, so that the resulting measurement sets obey $\mc{Y}\round{1} \subset \mc{Y}\round{2} \subset \cdots \subset \mc{Y}\round{C}$.
Then, for each configuration $k=1,\dots,C$, a calibration set $d\cal\round{k}$ is collected, from which the set-valued function $\mc{C}(\cdot;d\cal\round{k})$ is constructed.
At test time, we begin by collecting measurements $y\in\mc{Y}\round{1}$ according to the first (i.e., minimal) configuration.
From $y$ we compute the reconstructions $\{\hat{x}\of{j}\}_{j=1}^p$ and, from them, the task uncertainty $|\mc{C}(\{\hat{x}\of{j}\};d\cal\round{1})|$.
If this uncertainty falls below the desired $\tau$, we stop collecting measurements.
If not, we would collect the additional measurements in $\mc{Y}\round{2} \setminus \mc{Y}\round{1}$, and repeat the procedure.
\Cref{fig:multi_round_sampling} summarizes the proposed multi-round protocol.

\section{Numerical Experiments} \label{sec:experiments}

We now demonstrate our task-based uncertainty quantification framework on MRI \cite{Knoll:SPM:20}.
MRI offers exceptional soft-tissue contrast without ionizing radiation but suffers from very slow scan times.
Accelerated MRI speeds the acquisition process by collecting a fraction $1/R$ of the measurements specified by the Nyquist sampling theorem.
The integer $R$ is known as the ``acceleration rate.''
When $R>1$, the inverse problem is ill-posed.

In MRI, a typical task is to diagnose the presence or absence of a pathology.
Although this task is typically performed by a radiologist, neural-network-based classification is expected to play a significant role in aiding radiologists \cite{Boeken:DII:23}.
Thus, in our experiments, we implement the task $f(\cdot)$ using a neural network.
Details are given below.

\textit{Data:}~We use the multi-coil fastMRI knee dataset \cite{Zbontar_etal:18} and in particular the non-fat-suppressed subset, which includes 484 training volumes (17286 training slices, or images) and 100 validation volumes (2188 validation images).
We use pathology labels from fastMRI+ \cite{Zhao:SD:22}.
For knee-MRI, meniscus tears yield the largest fastMRI+ label set, and so we choose meniscus-tear-detection as our task.
To collect measurements, we retrospectively subsample the fastMRI data in the k-space using a set of random nested masks that yield acceleration rates $R\in\{16,8,4,2\}$, the details of which are described in the Supplementary Materials.

\textit{Image Recovery:}~We consider two recovery networks $g(\cdot)$.
As a point estimator, we use the state-of-the-art E2E-VarNet from \cite{Sriram:MICCAI:20} and,
as a posterior sampler, we use the conditional normalizing flow (CNF) from \cite{Wen:ICML:23}.
Both were specifically designed around the fastMRI dataset.
Another option would be the MRI diffusion sampler \cite{Chung:MIA:22}, but its performance is a bit worse than the CNF and its sampling speed is 8000$\times$ slower \cite{Wen:ICML:23}.
The E2E-VarNet and CNF were each trained to handle all four acceleration rates with a single model.

\textit{Task Network:}~We used a ResNet50 \cite{He:CVPR:16} for the task network $f(\cdot)$. 
Starting from an ImageNet-based initialization, we pretrained the weights to minimize the unsupervised SimCLR loss \cite{Chen:ICML:20} and later minimized binary-cross-entropy loss using the fastMRI+ labels.
See the Supplementary Materials for details.

\textit{Empirical Validation:}~Recall that the marginal coverage guarantee \eqref{mri_coverage} holds on average over random test samples $(\hat{X},Z)$ and random calibration data $D\cal=\{(\hat{X}_1,Z_1),\dots,(\hat{X}_n,Z_n)\}$.
To empirically validate marginal coverage and evaluate other average-performance metrics, we perform Monte-Carlo averaging over $T=10000$ trials as follows.
In each trial $t$, we randomly partition the 2188-sample validation dataset into a 70\% calibration fold with indices $i\in \mc{I}\cal[t]$ and a 30\% test fold with indices $i\in \mc{I}\test[t]$, construct conformal predictors using the calibration data $d\cal[t]=\{(\{\hat{x}_i\of{j}\}_{j=1}^p,z_i)\}_{i\in \mc{I}\cal[t]}$, and evaluate performance on test fold $t$.
Finally, we average performance over the $T$ trials.
Further details are given below.

\subsection{Effect of Acceleration Rate and Conformal Prediction Scheme} 

We have seen that the interval length $|\mc{C}(\{\hat{x}\of{j}\};d\cal)|$ provides a way to quantify the uncertainty that the measurement-and-reconstruction scheme contributes to the meniscus-classification task.
So a natural question is: How is the interval length affected by the MRI acceleration $R$?
We study this question below.

For a fixed acceleration $R$, the interval length is also affected by the choice of conformal predictor.
All else being equal, better conformal predictors yield smaller uncertainty sets  \cite{Angelopoulos:FTML:23}. 
So another question is: How is the interval length affected by selecting among the AR, LWR, or CQR conformal methods?

\begin{figure}[t]
  \centering
  \begin{subfigure}{0.49\linewidth}
    \includegraphics[width=1.0\linewidth]{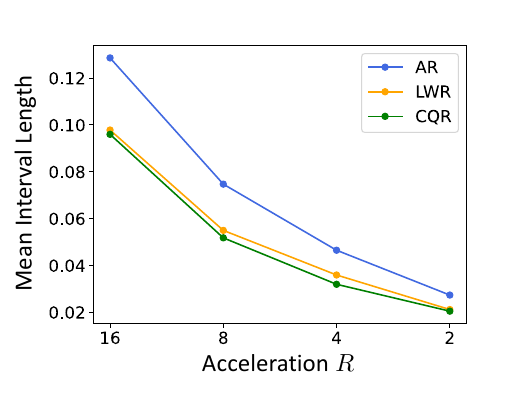}
    \caption{Mean Interval Length vs. $R$}
    \label{fig:mean_interval_v_R}
  \end{subfigure}
  \hfill
  \begin{subfigure}{0.49\linewidth}
    \includegraphics[width=1.0\linewidth]{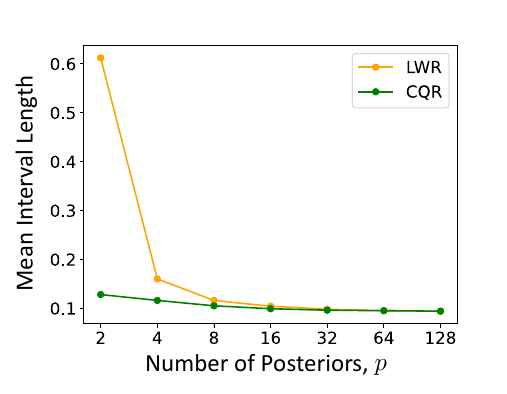}
    \caption{Mean Interval Length vs $p$}
    \label{fig:mean_interval_v_p}
  \end{subfigure}
  \caption{a) Average mean interval length versus acceleration $R$ with $p=32$ samples. b) Mean interval length versus $p$ with acceleration $R=16$. All results use error-rate $\alpha=0.05$ and $T=10000$ trials.  }
  \label{fig:mean_interval}
\end{figure}

%First, we analyze how our methods adapt the intervals to the difficulty of the input, otherwise referred to as local adaptivity \cite{Lei:JASA:18, Romano:NIPS:19, Romano:NIPS:20, Angelopoulos:ICLR:20}, by looking at the mean interval length.
To answer these questions, we compute the ``average mean interval length'' $\bar{\MIL}\defn\frac{1}{T}\sum_{t=1}^T \MIL[t]$ using the trial-$t$ mean interval length 
\begin{equation}
    %| \bar{\mathcal{C}}\of{t} | =
    \MIL[t] \defn
    \frac{1}{|\mc{I}\test[t]|} \sum_{i \in \mc{I}\test[t]}  
    \big| \mathcal{C}\big(\{\hat{x}_i\of{j}\}_{j=1}^p;d\cal[t]\big) \big| ,
\end{equation}
\Cref{fig:mean_interval} plots the average mean interval length versus $R$ for the AR, LWR, and CQR conformal predictors using $T=10000$ trials, $p=32$ posterior samples, and error-rate $\alpha=0.05$.
The figure shows that, as expected, the average mean interval length decreases as more measurements are collected (i.e., as $R$ decreases).
%As expected, the mean interval length increases with the acceleration rate since the inherent variability of the reconstructions increases.
The figure also shows that, as expected, the (locally adaptive) LWR and CQR methods give consistently smaller average mean interval lengths than the (non-adaptive) AR method.
%Both adaptive methods give consistently smaller mean interval lengths compared to the absolute residual, which suggests the adaptive nature allows them to provide tighter bounds on the ground truth soft output while providing the same coverage.
%In other words, the adaptive methods are able to provide smaller intervals for inputs with less uncertainty while the fixed-interval method can only provide an averaged interval that balances coverage for all types of inputs. 
In this sense, posterior sampling is advantageous over point sampling.

\subsection{Effect of Number of Posterior Samples}

Above, we saw that the measurement process and conformal method both affect the prediction-interval length.
We conjecture that the image reconstruction process will also affect the prediction-interval length.
To investigate this, we vary the number of samples $p$ produced by the posterior-sampling scheme, reasoning that smaller values of $p$ correspond to less accurate recoveries (e.g., a less accurate posterior mean approximation).

\Cref{fig:mean_interval_v_p} plots the average mean interval length versus $p$ for the LWR and CQR conformal predictors using $T=10000$ trials, acceleration $R=16$, and error-rate $\alpha=0.05$.
As expected, the interval length decreases as the posterior sample size $p$ grows.
But interestingly, LWR is much more sensitive to small values of $p$ than CQR.
One implication is that small values of $p$ may suffice when used with an appropriate conformal prediction method.

%For the adaptive methods, we look at the effect of the number of posterior samples $p$ used in the computation of the conformal interval. While generating many samples gives better estimates of the standard deviation and quantiles for the posterior soft outputs, it requires more compute and thus more time.

%In \cref{fig:mean_interval_v_p}, we plot the average mean interval length for $T=10000$ trials using different values of $p$, $R=16$, and $\alpha=0.05$. CQR performs much more consistently for all values of $p$ while LWR is very sensitive to low numbers of posteriors samples. CQR maintains a respectable mean interval length even with just two posterior samples, which could indicate a higher potential for extremely time-sensitive use cases.  Both methods saturate around $p=16$, suggesting beyond that one does not see much of an improvement in mean interval length relative to the increase in processing time.

\subsection{Empirical Validation of Coverage} \label{sec:coverage}

To verify that the marginal coverage guarantee \eqref{mri_coverage} holds, we compute the empirical coverage of Monte-Carlo trial $t$ as
%after the final measurement round using $T=10000$ Monte-Carlo trials:
\begin{align}
%\bar{\EC}
%\defn \frac{1}{T}\sum_{t=1}^T \EC[t]
%\quad\text{for}\quad
\EC[t] 
\defn \frac{1}{|\mc{I}\test[t]|} \sum_{i \in \mc{I}\test[t]} \mathds{1}\{z_i \in  \mathcal{C}(\{\hat{x}_i\of{j}\};d\cal[t])\} 
\label{eq:ECt},
\end{align}
where $\mathds{1}\{\cdot\}$ denotes the indicator function.
%As reported in \cref{table:multi_round}, $\bar{\EC}$ is very close to the anticipated value of $1-\alpha=0.95$ for all conformal methods.
Existing theory (see, e.g., \cite{Angelopoulos:FTML:23}) says that when $(\{\hat{X}_i\of{j}\},Z_i)$ and $D\cal[t]$ in \eqref{ECt} are exchangeable pairs of random variables, $\EC[t]$ is random and distributed as
\begin{align}
\EC[t] \sim \BetaBin(n\test,n\cal + 1 - l\cal,l\cal)
\quad\text{for}\quad
l\cal \defn \lceil(n\cal+1)\alpha\rceil 
\label{eq:Beta},
\end{align}
where $n\test \defn |\mc{I}\test[t]|$ and $n\cal \defn |\mc{I}\cal[t]|$. 

\begin{figure}[t]
    \centering
    \includegraphics[width=1.0\linewidth]{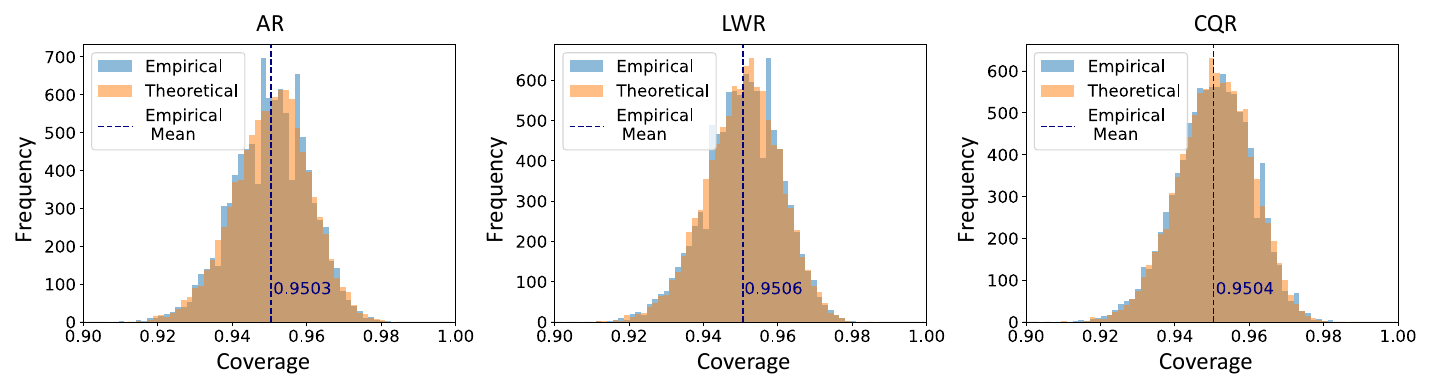}
    \caption{For the AR, LWR, and CQR conformal methods, each subplot shows the histograms of the empirical and theoretical empirical-coverage samples $\{\EC[t]\}_{t=1}^T$ across $T=10000$ Monte-Carlo trials using $\alpha=0.05$, $R=8$, and $p=32$. The subplots are also labelled with the empirical mean of $\{\EC[t]\}_{t=1}^T$, which is very close to the target value of $1-\alpha=0.95$.
    }
    \label{fig:distribution_validation}
\end{figure}

For each of the three conformal methods, \cref{fig:distribution_validation} shows the histogram of $\{\EC[t]\}_{t=1}^T$ from \eqref{ECt} for $T=10000$, error-rate $\alpha=0.05$, acceleration $R=8$, and $p=32$ posterior samples.
The figure shows that this histogram is close to the histogram created from $T$ samples of the theoretical distribution in \eqref{Beta}.
\Cref{fig:distribution_validation} also prints the average empirical coverage $\frac{1}{T}\sum_{t=1}^T \EC[t]$ for each method, which is very close to the target value of $1-\alpha=0.95$.
Thus we see that, in practice, conformal prediction behaves close to the theory.

\subsection{Multi-Round Measurements} \label{sec:multi_round_mri}

We now investigate the application of the multi-round measurement protocol from \cref{sec:protocol} to accelerated MRI.
For this, we simulated the collection of MRI slices over rounds $k=1,\dots,5$, stopping as soon as the $\alpha=0.01$ interval width $|\mc{C}(\{\hat{x}\of{j}\};d\cal\round{k})|$ falls below the threshold of $\tau=0.1$.
The first round collects k-space measurements at acceleration rate $R\round{1}=16$, and the remaining rounds each collect additional k-space measurements to yield $R\round{2}=8$, $R\round{3}=4$, $R\round{4}=2$, and $R\round{5}=1$ respectively.
For quantitative evaluation, we randomly selected 8 multi-slice volumes from the 100-volume fastMRI validation set to act as test volumes (half of which were labeled as meniscus tears and half of which were not), and we used the remaining 92 volumes for calibration.
We will refer to the corresponding index sets as $\mc{I}\test$ and $\mc{I}\cal$.
Additional details about the MRI measurement procedure are given in the Supplementary Materials.

We begin by discussing the AR conformal prediction method, which uses the point-sampling E2E-VarNet \cite{Sriram:MICCAI:20} for image recovery.
The AR method produces prediction intervals $\mc{C}(\hat{x};d\cal\round{k})$ that are $\hat{x}$-invariant (i.e., not locally adaptive).
Thus, immediately after calibration, it is known that $k=4$ measurement rounds (i.e., $R=2$) are necessary and sufficient to achieve the $\tau=0.1$ threshold
at error-rate $\alpha=0.01$.
%Note that this finding agrees with \cref{fig:mean_interval_v_R}, which plots the interval length versus $R$.

\begin{figure}[t]
    \centering
    \begin{minipage}{0.44\linewidth}
        \centering
        \includegraphics[width=1.0\linewidth,trim= 0 10 0 0,clip]{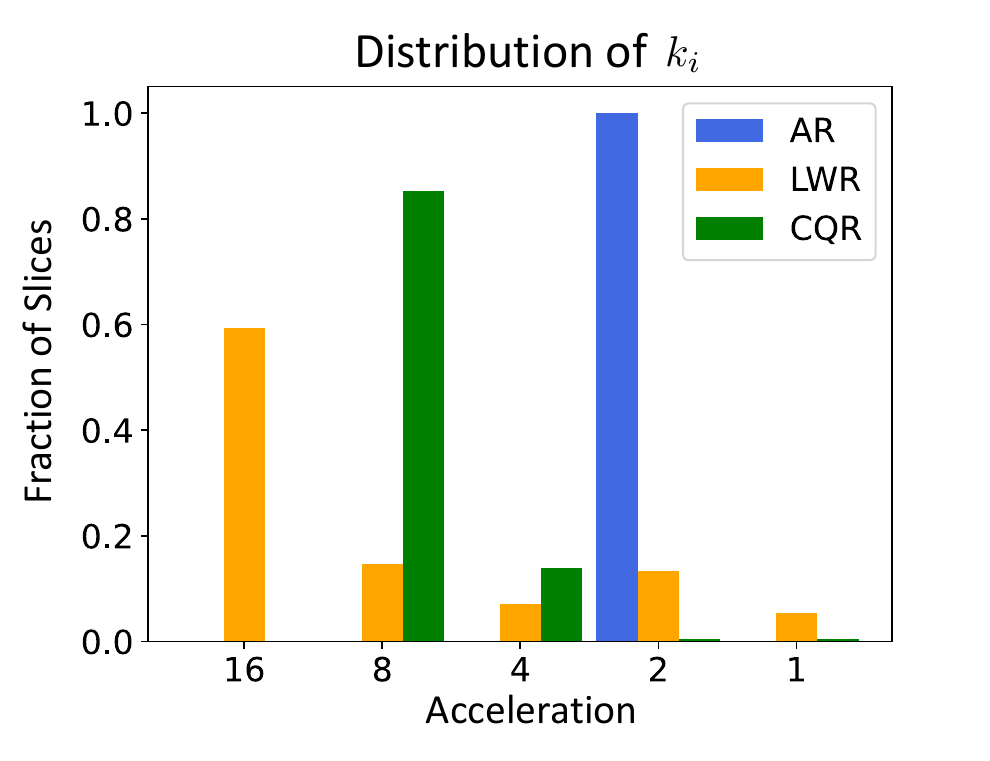}
        %\vspace{-6mm}
        \caption{Fraction of slices accepted after a given acceleration rate.}
        \label{fig:slices_remaining}
    \end{minipage}
    \hfill
    \begin{minipage}{0.55\linewidth}
        \centering
        \captionof{table}{Average metrics for the multi-round MRI simulation ($\pm$ standard error).
        }
        \resizebox{\columnwidth}{!}{%
        \begin{tabular}{|c|c|c|c|}
            \hline
            Method 
            & \multicolumn{1}{|p{2cm}|}{\centering Average \\ Acceleration}  
            & \multicolumn{1}{|p{2cm}|}{\centering Empirical \\ Coverage} 
            & \multicolumn{1}{|p{2cm}|}{\centering Average Max \\ Center Error} \\
            \hline
            AR & $2.000$ & $0.991 \pm 0.008$ & $0.032 \pm 0.017$ \\
            LWR & $5.157$ & $0.992 \pm 0.005$ & $0.020 \pm 0.002$ \\
            CQR & $6.762$ & $0.987 \pm 0.008$ & $0.044 \pm 0.009$ \\
            \hline
        \end{tabular}
        }
        \label{table:multi_round}
    \end{minipage}
\end{figure}

The LWR and CQR conformal prediction methods both use the CNF \cite{Wen:ICML:23} with $p=32$ posterior samples and
yield locally adaptive prediction intervals $\mc{C}(\{\hat{x}\of{j}\};d\cal\round{k})$.
This allows them to evaluate the interval length for each $\{\hat{x}\of{j}\}$ and stop the measurement process as soon as that length falls below the threshold $\tau$.
For test image $i\in\mc{I}\test$, we denote the final measurement round as
\begin{align}
k_i \defn \min \big\{k: |\mc{C}(\{\hat{x}_i\of{j}\};d\cal\round{k})| < \tau \big\} .
\end{align}
(Note that $\{\hat{x}_i\of{j}\}$ also changes with the measurement round $k$, although the notation does not explicitly show this.)
The average acceleration is then
\begin{align}
\bar{R} = \bigg( \frac{1}{|\mc{I}\test|} \sum_{i\in\mc{I}\test} \frac{1}{R\round{k_i}} \bigg)^{-1}.
\end{align}
\Cref{table:multi_round} shows the average acceleration $\bar{R}$ for the AR, LWR, and CQR conformal methods.
We see that $\bar{R}=2$ for the AR method because it always uses four measurement rounds.
The LWR and CQR methods achieve higher average accelerations $\bar{R}$ because fewer measurement rounds suffice in a large fraction of cases.
\Cref{table:multi_round} also shows that the empirical coverage is close to what we would expect given this relatively small test set.

\Cref{fig:slices_remaining} plots the distribution of final-round $\{k_i\}_{i\in\mc{I}\test}$ for the AR, LWR, and CQR conformal methods.
It too shows that the AR method always uses four rounds (i.e., $R=2$), while the LWR and CQR methods typically use fewer rounds.
However, this plot also shows that the LWR method sometimes uses five measurement rounds.
This may seem counter-intuitive but can be explained as follows.
At $k=4$, the AR method is calibrated so that the true score $z$ lands in the prediction interval in all but $\alpha=1\%$ of the cases, where the length of that interval is small enough to meet the $\tau=0.1$ threshold.  
%But the remaining $1\%$ of the cases can be arbitrarily bad.
Meanwhile, the LWR (and CQR) methods adapt the prediction interval based on the difficulty of $\{\hat{x}\of{j}\}$.
In most cases, the LWR prediction interval is smaller than the AR interval, but for a few ``difficult'' cases the LWR prediction interval is wider, and in fact too wide to meet the $\tau=0.1$ threshold. 
For these difficult cases, the LWR method moves on to the fifth measurement round.

Based on the previous discussion, one might conjecture that the prediction intervals accepted by the AR method at round $k=4$ will be somehow worse than those accepted by LWR at $k=4$, even though their lengths all meet the threshold.
We can confirm this by interpreting the midpoint of the prediction interval as an estimate of $z$ and evaluating the absolute error on that estimate, which we call the ``center error'' (CE):
\begin{equation}
\CE(\{\hat{x}\of{j}\},z) \defn \bigg| z - \frac{b_l+b_u}{2}  \bigg|
\quad\text{where}\quad
[b_l,b_u] = \mathcal{C}(\{\hat{x}\of{j}\};d\cal) .
\end{equation}
When evaluating the center error, we take the maximum over the slices in each volume.
\Cref{table:multi_round} lists the average maximum center error and confirms that it is smaller for LWR than for AR.

\begin{figure}[t]
    \centering
    \includegraphics[width=1\linewidth]{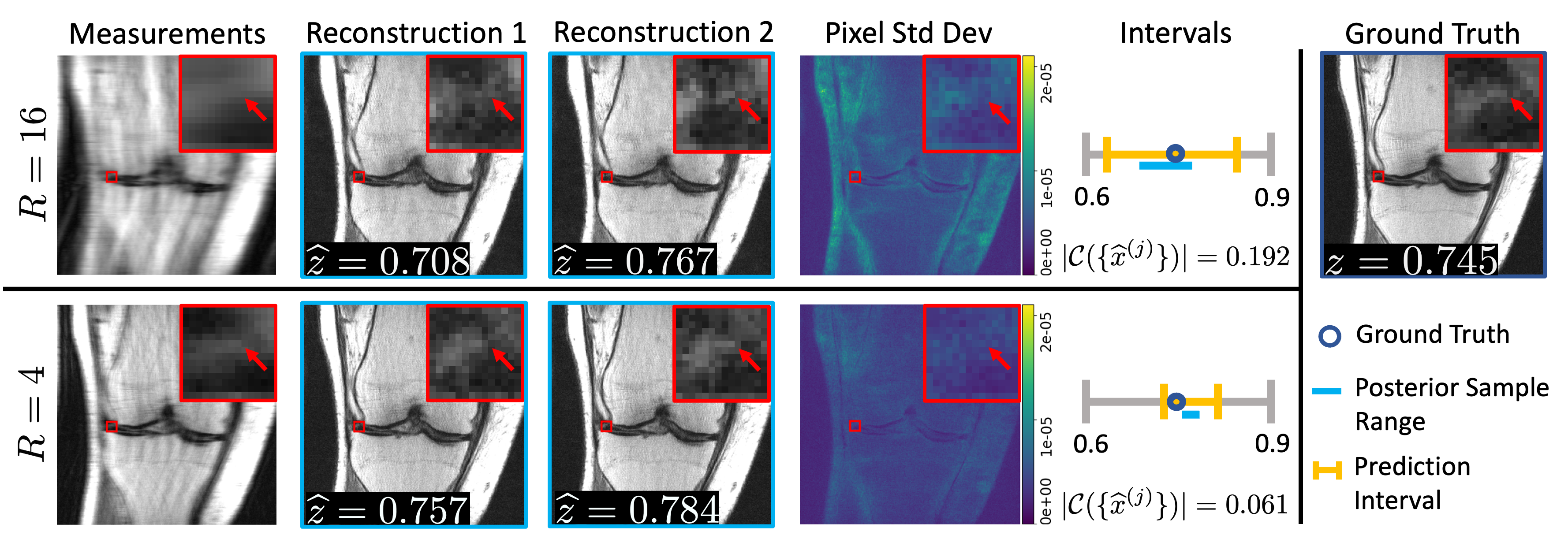}
    \caption{MR Image reconstructions and CQR prediction intervals at accelerations $R=16$ and $R=4$ with error-rate $\alpha=0.01$ and a total of $p=32$ posterior samples. The fastMRI+ bounding box around the meniscus tear is magnified in red. The prediction intervals shrink as the posterior samples become more consistent in the meniscus region. The standard-deviation maps show areas of high pixel-wise uncertainty but are difficult to connect to the downstream task. Note, image brightness was increased to better highlight the tear. Best viewed when zoomed.}
    \label{fig:mri_intervals}
\end{figure}

%Since it is non-adaptive, the absolute residual interval drops below the threshold for all slices at the same acceleration 8. The adaptive methods conclude less relevant or more output-conclusive slice scans very quickly allowing them to achieve an average slice acceleration over 13. All methods are close to the expected coverage, demonstrating how we can achieve high acceleration and still meet statistical guarantees on the downstream task. Note that small deviations in the coverage are expected due to the evaluation on a single calibration and test set. Moreover, LWR achieves a max center error less than $\tau/2=0.05$ and thus a nearly perfect threshold coverage.  This indicates that all ground truth predictions are within the $\mathcal{C}_{\tau}(y)$ interval at the time when measurements were no longer collected. By satisfying this requirement, LWR provides a robust option for ensuring the reference soft output is within a $\tau$ sized interval.  For the absolute residual and CQR, this could mean that certain slices are outside of $\mathcal{C}_{\tau}(y)$ some small percentage of the time.

\Cref{fig:mri_intervals} shows examples of image reconstructions, pixel-wise standard deviation maps, and CQR prediction intervals for a test image labeled with a meniscus tear. 
At higher accelerations like $R=16$, relatively large variations across posterior samples $\{\hat{x}\of{j}\}$ result in relatively large variations across classifier outputs $\{\hat{z}\of{j}\}$, which result in a large prediction interval, i.e., high uncertainty about the ground-truth classifier output $z$. 
At lower accelerations like $R=4$, relatively small variations across posterior samples yield smaller prediction intervals, i.e., less uncertainty about $z$.
While the pixel-wise standard-deviation maps also show reduced variation across posterior samples, it's difficult to draw conclusions about uncertainty in the downstream task from them.
For example, the same pixel-wise variations could result from a set of reconstructions that show clear evidence for a tear in some cases and clear evidence to the contrary in others, or from a set of reconstructions that show clear evidence for a tear in all cases but are corrupted by different noise realizations. 
Our uncertainty quantification methodology circumvents these issues by focusing on the task itself. 
Furthermore, by leveraging the framework of conformal prediction, it ensures that the uncertainty estimates are statistically meaningful.

As far as practical implementation is concerned, for each slice in a volume, the CNF reconstructions, ResNet-50 classifier outputs, and conformal prediction intervals can be computed in 414 milliseconds for $p=32$ samples, or 7.4 milliseconds for $p=2$ samples, on a single NVIDIA A100 GPU.

%\begin{table}[t]
%    \centering
%    \caption{Average metrics for the multi-round MRI simulation. ($\pm$ standard error)}
%    \begin{tabular}{|c|c|c|c|c|}
%        \hline
%        Method 
%        & \multicolumn{1}{|p{2.5cm}|}{\centering Average \\ Acceleration $\uparrow$}  
%        & \multicolumn{1}{|p{2.5cm}|}{\centering Empirical \\ Coverage $\uparrow$}
%        & \multicolumn{1}{|p{2.5cm}|}{\centering Max Center \\ Error $\downarrow$} 
%        & \multicolumn{1}{|p{2.5cm}|}{\centering Threshold \\ Coverage $\uparrow$} \\
%        \hline
%        AR & $8.000 \pm 0.000$ & $0.959 \pm 0.013$ & $0.064 \pm 0.015$ &  $0.979 \pm 0.008$  \\
%        LWR & $13.343 \pm 0.287$ & $0.945 \pm 0.018$ & $0.036 \pm 0.004$ & $0.997 \pm 0.003$  \\
%        CQR & $14.250 \pm 0.205$ & $0.951 \pm 0.017$ & $0.060 \pm 0.007$ & $0.966 \pm 0.015$  \\
%        \hline
%    \end{tabular}
%\end{table}

\section{Discussion} \label{sec:discussion}

A number of works on uncertainty quantification for MRI have been proposed based on Bayesian neural networks and posterior sampling, e.g., \cite{Schlemper:MLMIR:18,Edupuganti:TMI:20,Denker:JI:21,Jalal:NIPS:21,Ekmekci:TCI:22,Narnhofer:TMI:22,Chung:MIA:22,Wen:ICML:23,Bendel:NIPS:23}.
They produce a set of possible reconstructions $\{\hat{x}\of{j}\}$, from which a pixel-wise uncertainty map is typically computed.
Conformal prediction methods have also been proposed to generate pixel-wise uncertainty maps for MRI and other imaging inverse problems \cite{Angelopoulos:ICML:22,Kutiel:ICLR:23,Teneggi:ICML:23,Horwitz:22}, but with statistical guarantees.
However, when imaging is performed with the eventual goal of performing a downstream task, pixel-wise uncertainty maps are of questionable value.
In this work, we construct a conformal prediction interval that is statistically guaranteed to contain the task output from the true image. 
We focus on tasks that output a real-valued scalar, such as soft-output binary classification.
%and use conformal prediction to construct an interval $\mc{C}\subset\Real$ that is guaranteed to contain the ground-truth task output $z=f(x)$ with probability $1-\alpha$. The interval length $|\mc{C}|$ then quantifies the uncertainty that the measurement-and-reconstruction process contributes to the task.

Other works have applied conformal prediction to MRI tasks. 
Lu et al.\ \cite{Lu:MICCAI:22} consider a dataset $\{(x_i,z_i)\}$ with MRI images $x_i$ and discrete ordinal labels $z_i\in\{1,\dots,K\}$ that rate the severity of a pathology.
They design a predictor that, given test $x$, outputs a set $\mc{Z}(x)\in 2^K$ that is guaranteed to contain the true label $z$ with probability $1-\alpha$.
Different from our work, \cite{Lu:MICCAI:22} involves no inverse problem and aims to quantify the uncertainty in a discrete $z$.
Sankaranarayanan et al.\ \cite{Sankaranarayanan:NIPS:22} compute uncertainty intervals on the presence/absence of semantic attributes in images, and mention that one application could be pathology detection in MRI (although they do not pursue it).  
Although their high-level goal is similar to ours, their solution requires a trained ``disentangled'' generative network that, in the case of MRI, would generate MRI images from pathology probabilities.
To our knowledge, no such networks exist for MRI.
In contrast, our method requires only a trained pathology classifier $f(\cdot)$, which should be readily available.

\textbf{Limitations:} 
First, our method requires a downstream task, which is not always available.
Second, we demonstrated our method on only a single inverse problem and task;
validation on other applications is needed.
Third, our MRI application ideas are preliminary and not ready for clinical use.
Since we use the conformal prediction interval width as a proxy for the diagnostic value of the reconstructed image(s), several aspects of our design (e.g., the choice of classifier $f(\cdot)$, recovery algorithm $g(\cdot)$, conformal prediction method, threshold $\tau$, and error-rate $\alpha$) would need to be tuned and validated through rigorous clinical studies.
Fourth, for ease of exposition, the conformal methods that we use (AR, LWR, CQR) are somewhat simple. 
More advanced methods, like risk-controlling prediction sets (RCPS) \cite{Bates:JACM:21}, may perform better.
Fifth, we considered only tasks that output a single real-valued scalar, such as soft-output binary classification.
Extensions to more general tasks would be useful.
%We conjecture that our method could be extended to $K$-output tasks (e.g., multi-label classification) by constructing $K$ conformal prediction intervals $\mc{C}_k\subset\Real$ and quantifying uncertainty as the max interval-length, $\max_k |\mc{C}_k|$, but more work to verify this is needed.
Lastly, our posterior sampler only considers aleatoric uncertainty. 
In principle, epistemic uncertainty could be included by sampling the generator's weights from a distribution, as in \cite{Ekmekci:NIPSW:23}, but more work is needed in this direction.

\section{Conclusion}
For imaging inverse problems, we proposed a method to quantify how much uncertainty the measurement-and-reconstruction process contributes to a downstream task, such as soft-output classification. 
In particular, we use conformal prediction to construct an interval that is guaranteed to contain the task-output from the true image with high probability.
We showed that, with posterior-sampling-based image recovery methods, the prediction intervals can be made adaptive, and we proposed a multi-round measurement protocol that stops acquiring new data when the task uncertainty is sufficiently small. 
We applied our method to meniscus-tear detection in accelerated knee MRI and demonstrated significant gains in acceleration rate.
%As a broad framework, we hope to encourage further development in higher level UQ for understanding imaging inverse problems and methods.

%In this work, we present a novel framework for quantifying the downstream uncertainty contribution of a measurement process for a generic imaging inverse problem. 
%For a given inverse problem and downstream task, we would like to perform the downstream task just as well using the measurements as having access to the true image.
%Given the measurements from an inverse problem, our procedure outputs an interval that contains the downstream output of the true image with high probability. 
%Our method utilizes the principles of conformal prediction to provide statistical guarantees on how often the intervals must include the downstream output of the true image.
%The length of the intervals provide a scalar quantification of the uncertainty resulting from the measurement process and opens the door for multi-round sampling procedures.
%We demonstrate the merits of our method on accelerated MRI and meniscus tear classification.
%We illustrate a large reduction in the number of measurements collected for moslesst slices while bounding the downstream output of the ground truth image to a user-specified interval. 
%As our method serves as an early framework, we hope to encourage a larger interest in the direction of task-specific uncertainty quantification for imaging inverse problems. 

\section*{Acknowledgements}
This work was supported in part by the National Institutes of Health under Grant R01-EB029957.

\clearpage
\bibliographystyle{splncs04}
\bibliography{bibs/macros,bibs/machine,bibs/misc,bibs/mri,bibs/sparse,bibs/books,bibs/phase}

\clearpage
\appendix

\section{Conditional Coverage Experiments} \label{sec:conditional_coverage}

As described in \cref{sec:background}, conformal prediction typically provides only marginal coverage guarantees. 
%such as, for example 
%\begin{equation}
%\mathbb{P}\big(Z\in \mathcal{C}(\{\hat{X}\of{j}\};D\cal)\big|\{\hat{X}\of{j}\}=\{\hat{x}\of{j}\}\big) \geq 1 - \alpha
%\label{eq:cond_coverage}
%\end{equation}
%for every $\{\hat{x}\of{j}\}$. 
%In fact, complete conditional coverage is infeasible for most cases \cite{Vovk:ACML:12,Lei:JRSS:14}. 
%Therefore, it is important to stress test methods to see how they perform under different conditional cases.
%Ideally, coverage is held across all conditioning cases.
In this section, we empirically evaluate different forms of conditional coverage.
To do this, we use the Monte-Carlo testing procedure described in \cref{sec:experiments} with $T=10000$, $R=8$, $p=32$, and $\alpha=0.05$.

\begin{figure}[t]
  \centering
  \begin{subfigure}{0.49\linewidth}
    \includegraphics[width=1.0\linewidth]{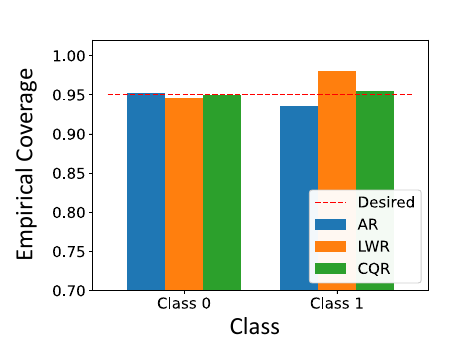}
    \caption{Class-Conditional Coverage}
    \label{fig:class_conditional_coverage}
  \end{subfigure}
  \hfill
  \begin{subfigure}{0.49\linewidth}
    \includegraphics[width=1.0\linewidth]{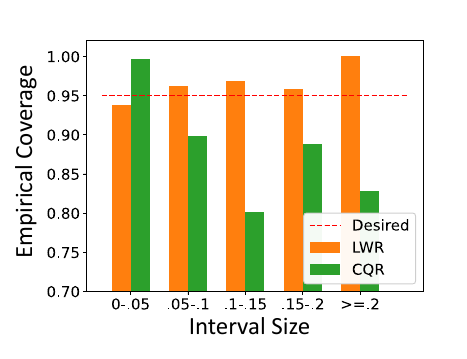}
    \caption{Size-Stratified Coverage}
    \label{fig:size_conditional_coverage}
  \end{subfigure}
  \caption{Coverage conditioned on class and interval-size for $T=10000$, $R=8$, $p=32$, and $\alpha=0.05$. Empirical coverage is close to $1-\alpha=0.95$ across both classes for each method. LWR maintains higher coverage across interval-sizes compared to CQR.}
  \label{fig:conditional coverage}
\end{figure}

First, we evaluate the class-conditional coverage 
\begin{equation}
\mathbb{P}\big(Z\in \mathcal{C}(\{\hat{X}\of{j}\};D\cal)\,\big|\,X \text{ labelled as $c$} \big)
%\label{eq:cond_coverage}
\end{equation}
for $c=0$ (no pathology) and $c=1$ (pathology).
%We evaluate the empirical coverage for all images of class 0 (no pathology) and class 1 (pathology) separately, analyzing if the coverage drops severely for either of the classes.
\Cref{fig:class_conditional_coverage} shows that, although all three methods provide close to the desired coverage of $95\%$, the AR method shows the worst under-coverage, which is $\approx 93\%$ for class 1.
Since a missed detection may carry higher consequence than a false alarm, under-coverage of class 1 should be considered carefully. 
Better class-conditional coverage could likely be attained using class-conditional conformal prediction \cite{Vovk:ACML:12}. 
%The same concept could be applied to provide equivalent coverage for other sensitive group splits like race, gender, or age.

Next we evaluate the size-stratified coverage \cite{Angelopoulos:ICLR:20}
\begin{equation}
\mathbb{P}\big(Z\in \mathcal{C}(\{\hat{X}\of{j}\};D\cal)\,\big|\,|\mathcal{C}(\{\hat{X}\of{j}\};D\cal)|\in\mc{S}\big) 
%\label{eq:cond_coverage}
\end{equation}
for size intervals $\mc{S} \in\{ [0,0.05], [0.05,0.1], [0.1,0.15], [0.15,0.2], [0.2,1]\}$.
%We break up the size of intervals into 5 bins spanning sizes of 0-0.05, 0.05-0.1, 0.1-0.15, 0.15-0.2, and $\geq$0.2. 
%We find the coverage for all images where the interval size falls within each of these bins, allowing us to see if the coverage holds well for all interval sizes.
\Cref{fig:size_conditional_coverage} shows that LWR demonstrates much more consistent size-stratified coverage than CQR.
%This may indicate that the CQR method lacks adequate adaptivity for harder inputs in our case, possibly a limitation from only utilizing an additive adjustment to the predicted quantiles. 
However, in cases such as multi-round sampling with $\tau\leq 0.05$, one may be concerned only with the coverage of small intervals, such as those with lengths $\leq 0.05$, where \cref{fig:size_conditional_coverage} suggests that CQR's coverage is very good. 
%Future improvements to our work should consider conditional coverage performance in evaluating readiness for real-world integration.  

\section{MRI Measurement Model}

Here we provide details on the accelerated multi-coil MRI setup used in our work.
Say that $w\in\Complex^l$ is a true 2D image that characterizes some slice of the human body.
Then a multi-coil MRI scanner produces spatial-Fourier-domain, or ``k-space,'' outputs 
\begin{equation}
u_b = P F S_b w + \varepsilon_b \in \Complex^m \text{~~for~~} b=1,\dots,B
\label{eq:kc} ,
\end{equation}
where $b$ is the coil index,
$P \in \Real^{m \times l}$ is a sampling matrix comprised of $m$ rows of the $l \times l$ identity matrix, 
$F \in \Complex^{l \times l}$ is the 2D unitary discrete Fourier transform (DFT) matrix,
$S_b \in \Complex^{l \times l}$ is a diagonal matrix that models the sensitivity of the $b$th coil,
and $\varepsilon_b \in \Complex^m$ is measurement noise \cite{Prussmann:MRM:99}.
A fully sampled MRI scan would use $m=l$, while an accelerated MRI scan would use $m < l$, where $R\defn l/m$ denotes the acceleration rate. 
%$R>1$ makes the inverse problem ill-posed.
%We will assume that $\{\vec{S}_c\}_{c=1}^C$ have been obtained from ESPIRiT \cite{Uecker:MRM:14}, in which case $\sum_{c=1}^C \vec{S}_c\herm\vec{S}_c=\vec{I}$.

Because the true coil maps $\{S_b\}$ are unknown and difficult to estimate, the MRI inverse problem can be formulated as recovering the ``coil images'' $x \defn [x_1\tran,\dots,x_B\tran]\tran$, where $x_b\defn S_b w$, instead of the true image $w$.
Given estimated coil images $\{\hat{x}_b\}$, the true-image magnitude $|w|$ can then be estimated using ``root-sum-of-squares'' (RSS) \cite{Roemer:MRM:90} as
\begin{align}
%\text{RSS}(\hat{x}) 
\hat{|w|}= \textstyle \sqrt{\sum_{b=1}^B |\hat{x}_b|^2}
\label{eq:rss}.
\end{align}
%We now simplify the forward process to take the form of a general linear inverse problem. 
%First, we form ``coil images'' $x_b \defn S_b w$ and stack them along the coil dimension . 
Furthermore, it is common to convert the k-space measurements $u_b$ to the spatial domain as $y_b \defn F\herm P \tran u_b$, which are called ``zero-filled'' images, and to stack them into $y \defn [y_1\tran,\dots,y_B\tran]\tran$ before feeding them to a recovery method.
Likewise, the stacked spatial-domain noise becomes $\epsilon = [(F\herm P \tran \varepsilon_1)\tran,\dots,(F \herm P \tran \varepsilon_B)\tran]\tran$.
Then, using 
\begin{align}
A \defn \blkdiag\big\{F \herm P \tran PF, \dots, F \herm P \tran PF \big\}
\label{eq:A},
\end{align}
we can write the inverse problem as described in \cref{sec:intro}:
\begin{align}
y
&= Ax + \epsilon .
\end{align}

%For context of this paper, we perform image recovery by estimating a multi-coil image $\hat{x}$ from the multi-coil zero-filled image $y$.
%Before being passed to the classifier, we combine the coils using the root-sum-of-squares (RSS)
%\begin{align}
%\text{RSS}(\hat{x}) = \textstyle \sqrt{\sum_{b=1}^B |\hat{x}_b|^2}
%\label{eq:rss}.
%\end{align}
%which provides a magnitude image estimate. 
%We repeat the magnitude image to form 3 channels before passing the estimate through the ResNet50.

\section{Mask Details}
%In \eqref{A}, the diagonal matrix $P\tran P$ has diagonal elements in $\{0,1\}$ and thus is said to perform a ``masking'' procedure in the spatial Fourier domain.
In \cref{sec:multi_round_mri}, we simulate a multi-round measurement process whereby, in round $k=1,\dots,5$, k-space samples are collected so that the accumulated samples up to round $k$ correspond to acceleration rate $R\round{k}\in \{16,8,4,2,1\}$.
The pattern of collected 2D samples is known as the ``sampling mask,'' 
%The subset of accumulated samples is often expressed as the vector $\diag(P\tran P)$, which has elements in $\{0,1\}$ and is known as the ``mask.'' 
and the choice of the mask can have a great impact on recovery quality.

In our work, we use Cartesian masks, where the sampling pattern consists of entire lines in 2D k-space.
For round $k=1$, we use a Golden Ratio Offset (GRO) mask \cite{Joshi:22} designed for rate $R\round{1}=16$ with the ``$\alpha$'' and ``s'' parameters from \cite{Joshi:22} chosen as $s=15$ and $\alpha=8$ (not to be confused with the meanings of $\alpha$ and $s$ elsewhere in this paper).
This provides a densely sampled rectangle in the center of the k-space, known as the autocalibration signal (ACS) region, which is $9$ lines wide.

%To simulate the accelerated MRI measurement process in \cref{sec:multi_round_mri}, we utilize four fixed, Cartesian sampling masks $P\tran P$ that indicate which lines of k-space are collected during the measurement process. 
%A mask is created for each acceleration factor, $R \in \{16,8,4,2\}$, and each subsequent acceleration's mask builds upon the mask of the previous acceleration in order to simulate a multi-round acquisition. 
%Note that these are GRO-specific parameters, which we use to be consistent with \cite{Joshi:22}, but are distinct from variable definitions elsewhere in the paper. 
%This provides an autocalibration signal (ACS) region of 9 pixels.

\begin{figure}
    \centering
    \includegraphics[width=1\linewidth]{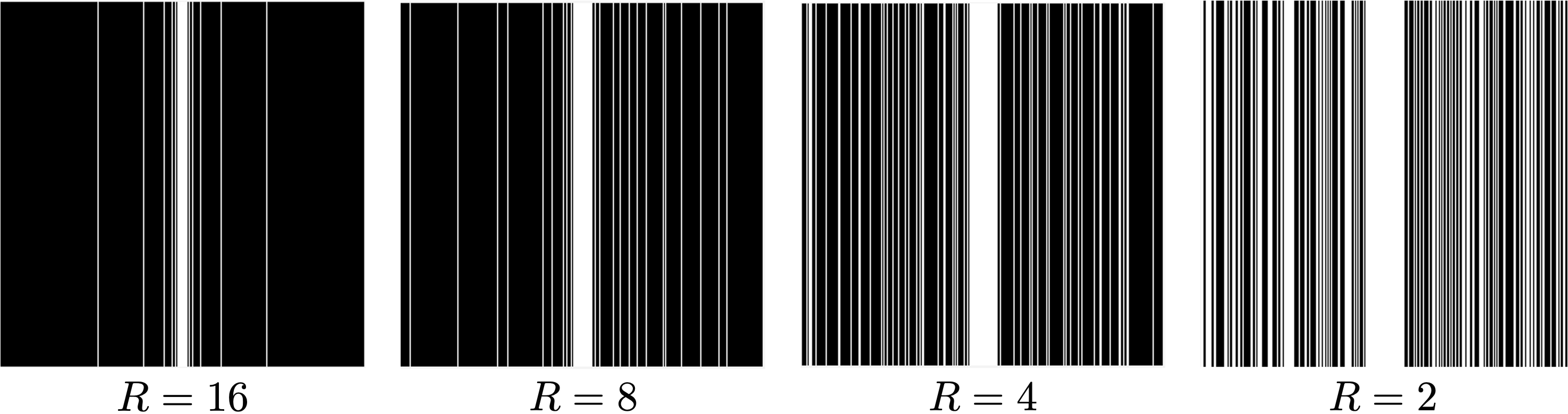}
    \caption{Sampling masks in the 2D spatial Fourier domain for each acceleration rate $R$. Each white line indicates the subset of collected samples. The masks are nested, in that the mask at a given $R$ contains all samples in all masks at higher $R$.}
    \label{fig:sampling_masks}
\end{figure}

For round $k=2$, we first collect $7$ additional lines in the center of k-space, yielding an ACS region of width $16$. 
We then sample additional k-space lines, using a sampling probability that is inversely proportional to the distance from the center, until an accumulated acceleration rate of $R\round{2}=8$ is achieved.
In the next two rounds, this procedure is repeated to obtain ACS regions of width $24$ and $32$, respectively, and accumulated accelerations of $R\round{3}=4$ and $R\round{4}=2$, respectively.
The four resulting masks are shown in \cref{fig:sampling_masks}.
The last round samples everywhere in k-space, achieving acceleration $R\round{5}=1$. 

\section{Data Details}

Here we provide further details about the datasets used in the paper. 
As mentioned in \cref{sec:experiments}, we use the non-fat suppressed subset of the multi-coil fastMRI knee dataset \cite{Zbontar_etal:18} for both the image recovery and pathology classification tasks. 
This contains 17286 training images from 484 training volumes.
From the\linebreak fastMRI+ \cite{Zhao:SD:22} annotations, there are 1921 images with meniscus tears.
For the validation set, we use the central 80\% of slice in a volume, which provides 2188 images.
Of these, 324 images contain meniscus tears.
For our Monte Carlo experiments, we randomly sample 70\% of the validation data for the calibration set and 30\% for the test set.
This results in 1531 calibration images and 656 testing images.
These dataset splits are summarized in \cref{table:dataset-splits}.

The fastMRI \cite{Zbontar_etal:18} dataset was collected by the NYU fastMRI initiative and is publicly available for research purpose.
Images and metadata have been deidentified and manually inspected for protected health information by the providers, and the curation of the dataset was part of an IRB-approved study.

\begin{table}[t]
\centering
\caption{Number of images in each data fold.}
\label{table:dataset-splits}
\begin{tabular}{|c|c|c|c|}
\hline
Training & \multicolumn{3}{c|}{Validation} \\ \cline{2-4}
               & Calibration & Testing & Total\\ \hline
17286 & 1531 & 656 & 2188 \\ \hline
\end{tabular}
\end{table}

\section{Network and Training Details}
For all models, we use an Adam optimizer \cite{Kingma:ICLR:15} with the default parameters, $\beta_{1} = 0.9$ and $\beta_{2}=0.999$.
We train each reconstruction network with all four accelerations.
More specifically, one of the four sampling masks is drawn uniformly at random for every sample in each epoch.
This allows the model to see each training sample at a different acceleration during training.

For the E2E VarNet, we utilize the author's implementation at \cite{Sriram:github:20}. 
We keep the default parameters listed for the fastMRI knee leaderboard and train to minimize the structural similarity (SSIM) \cite{Wang:TIP:04} loss for 50 epochs with a learning rate of $0.0001$ and batch size of $16$.

For the CNF, we modify the code from \cite{Wen:github:23}.
To better handle multiple accelerations, we increase the size of the conditioning network to have 256 initial channels. 
We also add an iMAP \cite{Sukthanker:CVPR:22} invertible attention module to the end of each flow step, and use 2 layers and 10 flow steps per layer.
The CNF is trained to minimize the negative log-likelihood for 150 epochs with a learning rate of $0.0001$ and batch size of $8$. 

For the soft-output binary classification network, we start with a ResNet50 \cite{He:CVPR:16} that is initialized with weights from a network trained on ImageNet \cite{Deng:CVPR:09}.
This task network takes in 3-channel images, so we convert the multi-coil image to a magnitude image using RSS \eqref{rss} and feed the magnitude image into all three input channels of the classifier. 
We pretrain the network with self-supervision using SimCLR \cite{Chen:ICML:20} for $500$ epochs with a learning rate of $0.0002$ and batch size of $128$. 
Next, we train the network in a supervised fashion to minimize the binary cross-entropy loss for $100$ epochs.
During the supervised training, we use an $\ell_2$-bounded projected gradient descent attack with $10$ steps and a perturbation budget of $1.5$ in order to make our network robust to $\ell_2$-bounded adversarial attacks.
The adversarial training is implemented using the robustness package \cite{robustness}.
The classifier is trained using a learning rate of $0.0001$ and batch size of $128$ with a weight decay of $0.01$. 
To prevent overfitting to the training data, we early-stop at the epoch that maximizes the area-under-the-receiver-operating-characteristic (AUROC) on the validation data.

All networks were implemented with PyTorch \cite{pytorch} and PyTorch Lightning \cite{lightning}.
Code is available at \url{https://github.com/jwen307/TaskUQ}.

\section{Image Recovery Performance}

Since our paper focuses on uncertainty metrics like prediction-interval length, the reader may wonder about how well the E2E-VarNet and CNF recovery approaches work according to traditional image-recovery metrics like peak-signal-to-noise ratio (PSNR), structural similarity index (SSIM) \cite{Wang:TIP:04}, and Fr\'echet Inception Score (FID) \cite{Heusel:NIPS:17}.
We provide those details in this section.

When computing these metrics, we used the RSS magnitude approximation from \eqref{rss}.
Also, following the approach of the fastMRI paper \cite{Zbontar_etal:18}, we compute PSNR and SSIM across entire volumes, rather than for each image slice separately.
For the CNF method, PSNR and SSIM are computed on the posterior-mean approximate computed by averaging $p=32$ posterior samples.
When computing FID, we use the VGG-16 embedding \cite{Simonyan:14} and $p=1$ samples for the CNF, and we compute reference statistics using the entire training set.

\Cref{table:reconstruction_metrics} shows the PSNR, SSIM, and FID performance of the E2E-VarNet and the CNF evaluated on the entire validation set at several accelerations $R$. 
There we see that the E2E-VarNet slightly outperforms the CNF in PSNR and SSIM, but not FID.
Because the models were trained to handle four different accelerations, the results in \cref{table:reconstruction_metrics} are slightly below those reported in the original E2E-VarNet and CNF papers \cite{Sriram:MICCAI:20} and \cite{Wen:ICML:23}. 

\begin{figure}
    \centering
    \includegraphics[width=1\linewidth]{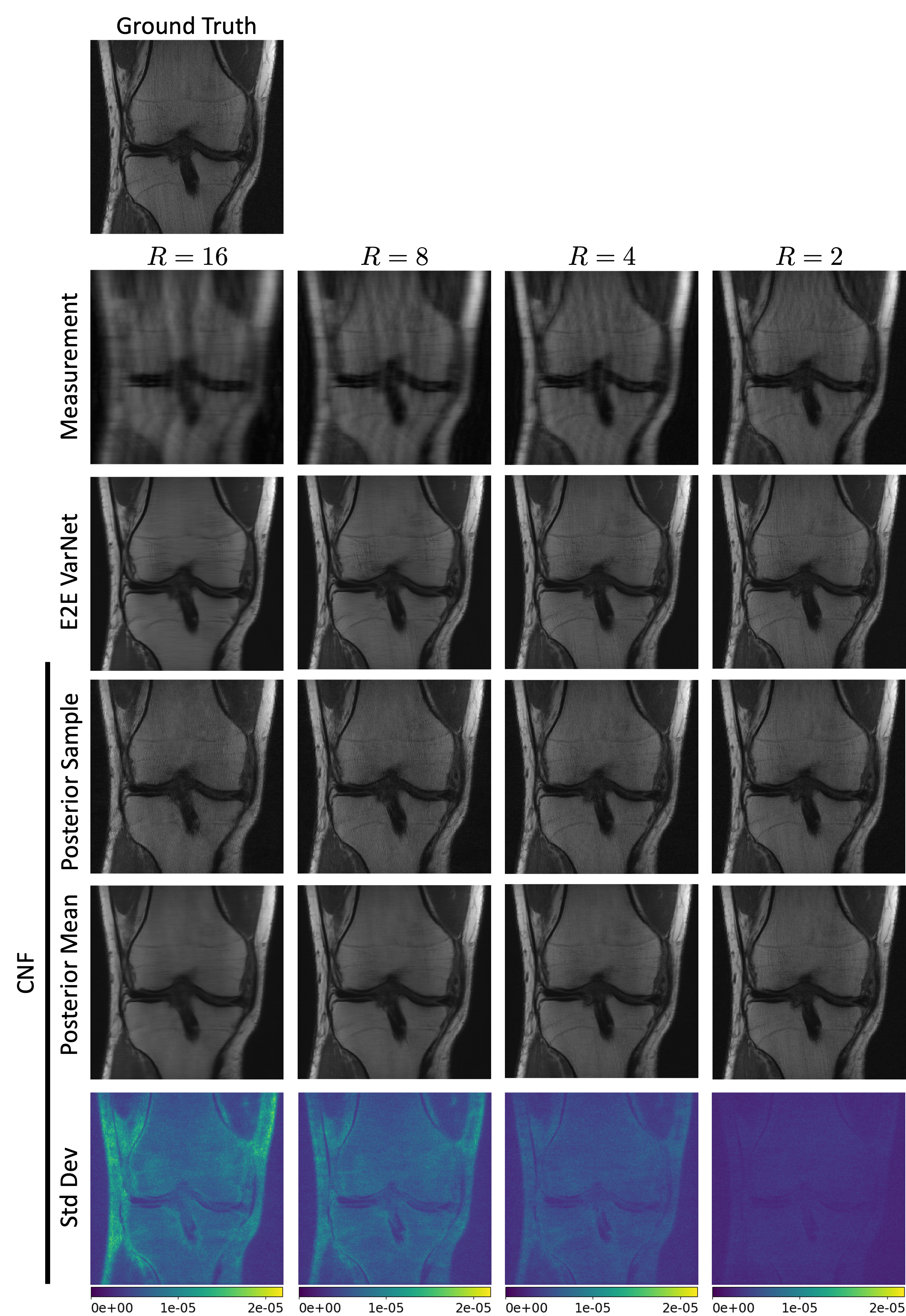}
    \caption{Example MRI reconstructions and standard-deviation maps for several accelerations $R$.}
    \label{fig:example_recons}
\end{figure}

\Cref{fig:example_recons} shows example reconstructions from the E2E-VarNet and CNF approaches, as well as standard-deviation maps for the CNF.
As expected, the posterior standard deviation decreases with the acceleration factor $R$. 

\begin{table*}[t]
\centering
\caption{Image-recovery metrics ($\pm$ standard error) versus acceleration $R$.}
\label{table:reconstruction_metrics}
\begin{tabular}{|c|c|p{2.5cm}|p{2.5cm}|p{2.5cm}|}
\hline
\multicolumn{1}{|c|}{$R$} & \multicolumn{1}{c|}{Network} & \multicolumn{1}{c|}{PSNR $\uparrow$} & \multicolumn{1}{c|}{SSIM $\uparrow$} & \multicolumn{1}{c|}{FID $\downarrow$} \\ \hline

\multirow{2}{*}{2} & E2E-VarNet & \multicolumn{1}{c|}{$42.341 \pm 0.273$} & \multicolumn{1}{c|}{$0.966 \pm 0.002$} & \multicolumn{1}{c|}{$2.847$} \\ \cline{2-5}
& CNF & \multicolumn{1}{c|}{$41.366 \pm 0.248$} & \multicolumn{1}{c|}{$0.960 \pm 0.002$} & \multicolumn{1}{c|}{$2.575$} \\ \cline{2-5} \hline

\multirow{2}{*}{4} & E2E-VarNet & \multicolumn{1}{c|}{$38.700 \pm 0.238$} & \multicolumn{1}{c|}{$0.937 \pm 0.003$} & \multicolumn{1}{c|}{$4.048$} \\ \cline{2-5}
& CNF & \multicolumn{1}{c|}{$37.974 \pm 0.216$} & \multicolumn{1}{c|}{$0.926 \pm 0.003$} & \multicolumn{1}{c|}{$3.349$} \\ \cline{2-5} \hline

\multirow{2}{*}{8} & E2E-VarNet & \multicolumn{1}{c|}{$36.120 \pm 0.212$} & \multicolumn{1}{c|}{$0.906 \pm 0.003$} &  \multicolumn{1}{c|}{$5.680$} \\ \cline{2-5}
& CNF & \multicolumn{1}{c|}{$35.593 \pm 0.196$} & \multicolumn{1}{c|}{$0.892 \pm 0.004$} & \multicolumn{1}{c|}{$4.346$} \\ \cline{2-5} \hline

\multirow{2}{*}{16} & E2E-VarNet & \multicolumn{1}{c|}{$32.911 \pm 0.194$} & \multicolumn{1}{c|}{$0.859 \pm 0.004$} & \multicolumn{1}{c|}{$9.668$} \\ \cline{2-5}
& CNF & \multicolumn{1}{c|}{$32.684 \pm 0.177$} & \multicolumn{1}{c|}{$0.842 \pm 0.004$} & \multicolumn{1}{c|}{$6.233$ } \\ \cline{2-5} \hline

\end{tabular}
\end{table*}

\section{Performance of Classifier}
The reader may also wonder about the performance of our soft-output meniscus-tear classifier according standard classification metrics. 
\Cref{table:classification_metrics} shows the accuracy, precision, recall, and AUROC evaluated on the validation set described in \cref{sec:experiments} with meniscus tear annotations from fastMRI+ \cite{Zhao:SD:22}.
From the table, we see that our classifier exhibits relatively high recall but low precision, which may be preferable in the context of meniscus-tear diagnosis, where missed detections might be more costly than false alarms. 
Given that the dataset used to train the classifier was relatively small, we conjecture that these performance metrics could be greatly improved with more data and a better balancing across classes.
That said, we believe that this classifier suffices as a task-based uncertainty evaluation tool.

\begin{table}
    \centering
    \caption{Validation performance of the meniscus-tear classifier}
    \begin{tabular}{|c|c|c|c|}
        \hline
        Accuracy & Precision & Recall & AUROC \\
        \hline
        0.775 &  0.391 & 0.929 & 0.922 \\
        \hline
    \end{tabular}
    \label{table:classification_metrics}
\end{table}

\end{document}